\documentclass[11pt,a4paper]{article}
\usepackage[hyperref]{naaclhlt2019}
\usepackage{times}
\usepackage{latexsym}
\usepackage{amsmath}
\usepackage{amssymb}
\usepackage{color,colortbl}
\usepackage{booktabs}
\usepackage{subcaption}
\usepackage{multirow}
\usepackage{url}
\usepackage{ifthen}
\usepackage{xspace}
\usepackage{graphicx}
\usepackage{textcomp}
\usepackage{float}
\usepackage{todonotes}
\usepackage{xspace}
\usepackage[utf8]{inputenc}
\aclfinalcopy 

\newif\ifcomment
\commentfalse

\newcommand\p[1]{\ensuremath{\left( #1 \right)}} 

\newcommand\eqdef{\ensuremath{\stackrel{\rm def}{=}}} 
\newcommand\refeqn[1]{(\ref{eqn:#1})}

\newcommand\refsec[1]{Section~\ref{sec:#1}}

\newcommand\reffig[1]{Figure~\ref{fig:#1}}

\newcommand\reftab[1]{Table~\ref{tab:#1}}

\ifthenelse{\isundefined{\definition}}{\newtheorem{definition}{Definition}}{}
\ifthenelse{\isundefined{\assumption}}{\newtheorem{assumption}{Assumption}}{}
\ifthenelse{\isundefined{\hypothesis}}{\newtheorem{hypothesis}{Hypothesis}}{}
\ifthenelse{\isundefined{\proposition}}{\newtheorem{proposition}{Proposition}}{}
\ifthenelse{\isundefined{\theorem}}{\newtheorem{theorem}{Theorem}}{}
\ifthenelse{\isundefined{\lemma}}{\newtheorem{lemma}{Lemma}}{}
\ifthenelse{\isundefined{\corollary}}{\newtheorem{corollary}{Corollary}}{}
\ifthenelse{\isundefined{\alg}}{\newtheorem{alg}{Algorithm}}{}
\ifthenelse{\isundefined{\example}}{\newtheorem{example}{Example}}{}

\newcommand{\retrieve}{\textsc{Retrieve}\xspace}
\newcommand{\retrieveSwap}{\textsc{Retrieve+Swap}\xspace}
\newcommand{\retrieveSwapTopic}{\textsc{Retrieve+Swap+Topic}\xspace}
\newcommand{\retrieveSwapTopicCombiner}{\textsc{Retrieve+Swap+Topic+Smoother}\xspace}

\newcommand{\pw}{\ensuremath{w^\text{p}}\xspace}
\newcommand{\aw}{\ensuremath{w^\text{a}}\xspace}

\newcommand{\sts}{sequence-to-sequence\xspace}

\newcommand{\ours}{\textsc{SemEval}\xspace}
\newcommand{\kao}{\textsc{Kao}\xspace}

\newcommand{\sratio}{\ensuremath{S_{\text{ratio}}}\xspace}

\newcommand{\pku}{\textsc{NeuralJointDecoder}\xspace}
\newcommand{\system}{\textsc{SurGen}\xspace}

\newcommand{\blank}{\underline{\hspace{1cm}}\xspace}

\newcolumntype{L}[1]{>{\raggedright\let\newline\\\arraybackslash\hspace{0pt}}m{#1}}
\newcolumntype{C}[1]{>{\centering\let\newline\\\arraybackslash\hspace{0pt}}m{#1}}
\newcolumntype{R}[1]{>{\raggedleft\let\newline\\\arraybackslash\hspace{0pt}}m{#1}}

\newcommand{\Note}[2]{} 
\newcommand{\SideNote}[2]{} 
\renewcommand{\Note}[2]{\todo[color=#1,size=\small, inline=true]{#2}} 
\renewcommand{\SideNote}[2]{\todo[color=#1,size=\small]{#2}} %
\newcommand{\NoteNP}[1]{\SideNote{purple!40}{#1 --Nanyun}}

\definecolor{darkgreen}{rgb}{0,0.5,0}
\ifcomment
\newcommand\pl[1]{\textcolor{red}{[PL: #1]}}
\newcommand\hh[1]{\textcolor{blue}{[HH: #1]}}
\newcommand\np[1]{\textcolor{darkgreen}{[NP: #1]}}
\else
\newcommand\pl[1]{}
\newcommand\hh[1]{}
\newcommand\np[1]{}
\renewcommand\NoteNP[1]{}
\fi

\newcommand\nl[1]{``\textit{#1}''}

\title{Pun Generation with Surprise}

\author{
    He He$^{1}$\thanks{$\;\;$Equal contribution.} \and Nanyun Peng$^{2}$\footnotemark[1] \and Percy Liang$^1$  \\
    $^1$Computer Science Department, Stanford University \\
    $^2$Information Sciences Institute, University of Southern California \\
    {\tt \{hehe,pliang\}@cs.stanford.edu, npeng@isi.edu}
}

\date{}

\begin{document}
\maketitle
\begin{abstract}
    We tackle the problem of generating a pun sentence given a pair of homophones (e.g., \nl{died} and \nl{dyed}).
Supervised text generation is inappropriate due to the lack of a large corpus of puns,
and even if such a corpus existed, mimicry is at odds with generating novel content.
In this paper, we propose an unsupervised approach to pun generation using a corpus of unhumorous text
and what we call the \emph{local-global surprisal principle}:
we posit that in a pun sentence,
there is a strong association between the pun word (e.g., \nl{dyed}) and the distant context,
as well as a strong association between the alternative word (e.g., \nl{died}) and the immediate context.
This contrast creates surprise and thus humor.
We instantiate this principle for pun generation in two ways:
(i) as a measure based on the ratio of probabilities under a language model,
and (ii) a retrieve-and-edit approach based on words suggested by a skip-gram model.
Human evaluation shows that our retrieve-and-edit approach generates puns successfully
31\% of the time, tripling the success rate of a neural generation baseline.

\end{abstract}

\section{Introduction}
\label{sec:intro}
\begin{figure}
\centering
\includegraphics[scale=0.45]{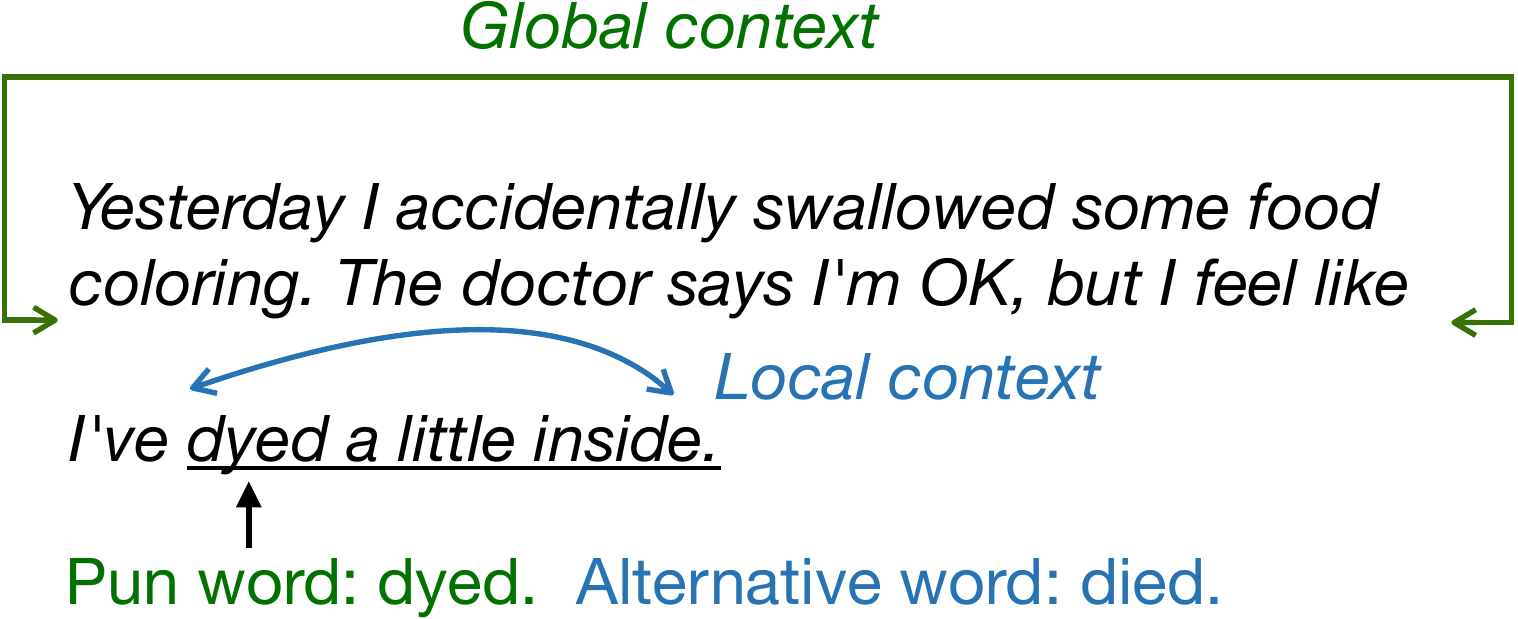}
\caption{An illustration of a homophonic pun.
  The \emph{pun word} appears in the sentence, while the \emph{alternative word}, which has the same pronunciation but different meaning, is implicated.
  The \emph{local context} refers to the immediate words around the pun word,
  whereas the \emph{global context} refers to the whole sentence.
}
\label{fig:pun_example}
\vspace{-1em}
\end{figure}

Generating \emph{creative} content is a key requirement in many natural language generation tasks
such as poetry generation~\cite{manurung2000towards,ghazvininejad2016poem}, story generation~\cite{meehan1977tale,peng2018towards,fan2018hierarchical,peng2019plan}, and social chatbots~\cite{weizenbaum1966eliza,fang2018sounding}.
In this paper, we explore creative generation with a focus on puns.
We follow the definition of puns in \newcite{aarons2017puns,miller2017semeval}: ``A pun is a form of wordplay in which one sign (e.g., a word or a phrase) suggests two or more meanings by exploiting polysemy, homonymy, or phonological similarity to another sign, for an intended humorous or rhetorical effect.''
We focus on a typical class of puns where the ambiguity comes from two (near) homophones. 
Consider the example in \reffig{pun_example}:
\nl{Yesterday I accidentally swallowed some food coloring. The doctor says I'm OK, but I feel like I've dyed (died) a little inside.}.
The \emph{pun word} shown in the sentence (\nl{dyed}) indicates one interpretation:
the person is colored inside by food coloring.
On the other hand, an \emph{alternative word} (\nl{died}) is implied by the context for another interpretation:
the person is sad due to the accident.

Current approaches to text generation require lots of training data,
but there is no large corpus of puns.
Even such a corpus existed,
learning the distribution of existing data and sampling from it is unlikely to lead to truly novel, creative sentences.
Creative composition requires deviating from the norm,
whereas standard generation approaches seek to mimic the norm.

Recently, \citet{yu2018neural} proposed an unsupervised approach that generates puns from a neural language model
by jointly decoding conditioned on both the pun and the alternative words,
thus injecting ambiguity to the output sentence.
However, \citet{kao2015pun} showed that ambiguity alone is insufficient to bring humor;
the two meanings must also be supported by distinct sets of words in the sentence.

Inspired by \citet{kao2015pun}, we propose a general principle for puns which we call \emph{local-global surprisal principle}.
Our key observation is that the strength for the interpretation of the pun and the alternative words flips as one reads the sentence.
For example, in \reffig{pun_example},
\nl{died} is favored by the immediate (local) context,
whereas \nl{dyed} is favored by the global context (i.e. \nl{...food coloring...}).
Our surprisal principle
posits that the pun word is much more surprising in the local context than in the global context,
while the opposite is true for the alternative word.

We instantiate our local-global surprisal principle in two ways.
First, we develop a quantitative metric for surprise based on the conditional probabilities of the pun word and the alternative word given local and global
contexts under a neural language model. 
However, we find that this metric is not sufficient for generation.
We then develop an unsupervised approach to generate puns based on a retrieve-and-edit framework~\cite{guu2018edit,hashimoto2018edit} given an unhumorous corpus (\reffig{overview}). 
We call our system \system (SURprisal-based pun GENeration).

We test our approach on 150 pun-alternative word pairs.\footnote{Our code and data are available at \url{https://github.com/hhexiy/pungen}.}
First, we show a strong correlation between our surprisal metric and funniness ratings from crowdworkers.
Second, human evaluation shows that our system 
generates puns successfully 31\% of the time, compared to 9\% of a neural generation baseline~\cite{yu2018neural},
and results in higher funniness scores.

\section{Problem Statement}
\label{sec:problem}

We assume access to a large corpus of raw (unhumorous) text.
Given a pun word \pw (e.g., \nl{dyed}) and an alternative word \aw (e.g., \nl{died}) which are (near) homophones,
we aim to generate a list of pun sentences.
A pun sentence contains only the pun word \pw,
but both \pw and \aw should be evoked by the sentence.

\section{Approach}
\label{sec:approach}

\subsection{Surprise in Puns}
\label{sec:surprisal}
What makes a good pun sentence?
Our key observation is that as a reader processes a sentence,
he or she expects to see the alternative word at the pun word position,
and are tickled by the relation between the pun word and the rest of the sentence.
Consider the following cloze test:
\nl{Yesterday I accidentally swallowed some food coloring. The doctor says I'm OK, but I feel like I've \blank a little inside.}.
Most people would expect the word in the blank to be \nl{died}
whereas the actual word is \nl{dyed}.
Locally, \nl{died a little inside} is much more likely than \nl{dyed a little inside}.
However, globally when looking back at the whole sentence,
\nl{dyed} is evoked by \nl{food coloring}.

Formally, \pw is more surprising relative to \aw in the local context,
but much less so in the global context.
We hypothesize that this contrast between local and global surprisal creates humor.

\subsection{A Local-Global Surprisal Measure}

Let us try to formalize the local-global surprisal principle quantitatively.
To measure the amount of surprise due to seeing the pun word instead of the alternative word in a certain context $c$,
we define surprisal $S$ as the log-likelihood ratio of the two events:
\begin{align}
\label{eqn:surprisal}
  S(c) \eqdef - \log \frac{p(\pw \mid c)}{p(\aw \mid c)} = - \log \frac{p(\pw, c)}{p(\aw, c)} .
\end{align}
We define the local surprisal to only consider context of a span around the pun word,
and the global surprisal to consider context of the whole sentence. 
Letting $x_1,\ldots, x_n$ be a sequence of tokens,
and $x_p$ be the pun word \pw,
we have
\begin{align}
\label{eqn:local_global}
    S_{\text{local}} &\eqdef S(x_{p-d:p-1}, x_{p+1:p+d}), \\
    S_{\text{global}} &\eqdef S(x_{1:p-1}, x_{p+1:n}),
\end{align}
where $d$ is the
local window size.

For puns, both the local and global surprisal should be positive because they are unusual sentences by nature.
However, the global surprisal should be lower than the local surprisal due to topic words hinting at the pun word.
We use the following unified metric, \emph{local-global surprisal}, to quantify whether a sentence is a pun:
\begin{align}
\label{eqn:ratio}
    S_{\text{ratio}} \eqdef
    \begin{cases}
        -1 & S_{\text{local}} < 0 \;\text{or}\; S_{\text{global}} < 0, \\
        S_{\text{local}} / S_{\text{global}} & \text{otherwise}.
    \end{cases}
\end{align}
We hypothesize that larger $S_\text{ratio}$ is indicative of a good pun.
Note that this hypothesis is invalid
when either $S_{\text{local}}$ or $S_{\text{global}}$ is negative,
in which case we consider the sentences equally unfunny by setting $S_\text{ratio}$ to $-1$.

\subsection{Generating Puns}
\label{sec:pungen}
The surprisal metric above can be used to assess whether a sentence is a pun,
but to generate puns, we need a procedure that can ensure grammaticality.
Recall that the surprisal principle requires
(1) a strong association between the alternative word and the local context;
(2) a strong association between the pun word and the distant context;
and (3) both words should be interpretable
given local and global context to maintain ambiguity.

Our strategy is to model puns as deviations from normality.
Specifically,
we mine \emph{seed sentences}
(sentences with the potential to be transformed into puns) from a large, generic corpus,
and edit them to satisfy the three requirements above.

\begin{figure}[t]
\centering
\includegraphics[scale=0.5]{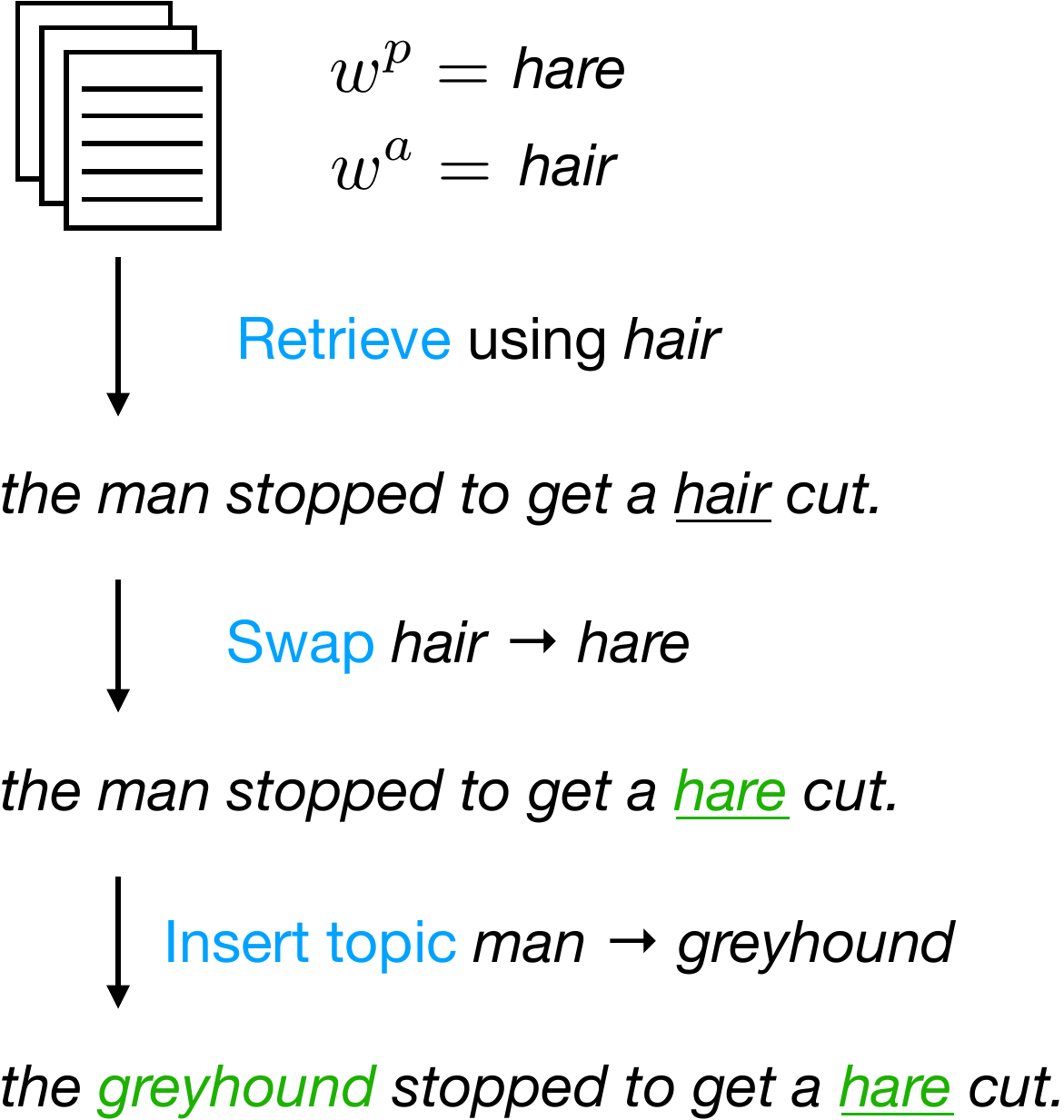}
\caption{Overview of our pun generation approach.
Given a pair of pun/alternative word,
we first retrieve sentences containing \aw from a generic corpus.
Next, \aw is replaced by \pw to increase local surprisal.
Lastly, we insert a topic word at the beginning of the sentence to create global associations supporting \pw and decrease global surprisal.
}
\label{fig:overview}
\end{figure}

\reffig{overview} gives an overview of our approach. Suppose we are generating a pun given $\pw = $ \nl{hare} and $\aw = $ \nl{hair}.
To reinforce $\aw =$ \nl{hair} in the local context despite the appearance of \nl{hare},
we retrieve sentences containing \nl{hair} and replace occurrences of it with \nl{hare}.
Here, the local context strongly favors the alternative word (\nl{hair cut}) relative to the pun word (\nl{hare cut}).
Next, to make the pun word \nl{hare} more plausible, we insert a \nl{hare}-related topic word (\nl{greyhound}) near the beginning of the sentence.
In summary, we create local surprisal by putting \pw in common contexts for \aw,
and connect \pw to a distant topic word by substitution. 
We describe each step in detail below.

\paragraph{Local surprisal.}
The first step is to retrieve sentences containing \aw.
A typical pattern of pun sentences is that the pun word only occurs once towards the end of the sentence,
which separates local context from pun-related topics at the beginning.
Therefore, we retrieve sentences containing exactly one \aw
and rank them by the position of \aw in the sentence (later is better).
Next, we replace \aw in the retrieved sentence with \pw.
The pun word usually fits in the context
as it often has the same part-of-speech tag as the alternative word.
Thus the swap creates local surprisal by putting the pun word in an unusual but acceptable context.
We call this step \retrieveSwap, and use it as a baseline to generate puns.

\paragraph{Global surprisal.}
While the pun word is locally unexpected,
we need to foreshadow it.
This global association must not be too strong
that it eliminates the ambiguity.
Therefore, we include a single \emph{topic word} related to the pun word by replacing one word at the beginning of the seed sentence.
We see this simple structure in many human-written puns as well.
For example, \nl{Old \underline{butchers} never die, they only \underline{meat} their fate.}, 
where pun words and their corresponding topic words are underlined.

We define relatedness between two words $w_i$ and $w_j$ based on a ``distant'' skip-gram model $p_\theta(w_j \mid w_i)$,
where we train $p_\theta$ to maximize $p_\theta(w_j \mid w_i)$ for all $w_i,w_j$ in the same sentence between $d_1$ to $d_2$ words apart.
Formally:
\begin{align}
\label{eqn:skipgram}
\sum_{j=i-d_1}^{i-d_2}\log p_\theta(w_j \mid w_i) +
\sum_{j=i+d_1}^{i+d_2}\log p_\theta(w_j \mid w_i).
\end{align}
We take the top-$k$ predictions from $p_\theta(w\mid\pw)$, where $\pw$ is the pun word, as candidate topic words $w$ to be further filtered next.

\paragraph{Type consistent constraint.}
The replacement must maintain acceptability of the sentence.
For example, changing \nl{person} to \nl{ship} in \nl{Each person must pay their fare share} does not make sense even though \nl{ship} and \nl{fare} are related.
Therefore, we restrict the deleted word in the seed sentence to nouns and pronouns,
as verbs have more constraints on their arguments and replacing them is likely to result in unacceptable sentences.

In addition, we select candidate topic words that are type-consistent with the deleted word, e.g., replacing \nl{person} with \nl{passenger} as opposed to \nl{ship}.
We define type-consistency (for nouns) based on WordNet path similarity.\footnote{
Path similarity is a score between 0 and 1 that is inversely proportional to the shortest distance between two word senses in WordNet.} 
Given two words,
we get their synsets from WordNet constrained by their POS tags.\footnote{
    Pronouns are mapped to the synset \texttt{person.n.01}.
}
If the path similarity between any pair of senses from the two respective synsets 
is larger than a threshold,
we consider the two words type-consistent.
In summary, the first noun or pronoun in the seed sentence is replaced by
a type-consistent topic word.
We call this baseline \retrieveSwapTopic.

\paragraph{Improve grammaticality.}
Directly replacing a word with the topic word may result in ungrammatical sentences,
e.g., replacing \nl{i} with \nl{negotiator} and getting
\nl{negotiator am just a woman trying to peace her life back together.}.
Therefore, we use a \sts model to smooth the edited sentence (\retrieveSwapTopicCombiner).

We smooth the sentence by deleting words around the topic word and train a model to fill in the blank.
The smoother is trained in a similar fashion to denoising autoencoders:
we delete immediate neighbors of a word in a sentence,
and ask the model to reconstruct the sentence by predicting missing neighbors. 
A training example is shown below:

\begin{table}[h]
    \newcommand{\sent}[1]{\small{\texttt{#1}}}
    \begin{tabular}{L{1.4cm}L{6cm}}
    Original: & \sent{the man slowly walked towards the woods .} \\
    Input: & \sent{<i> man </i> walked towards the woods .} \\
    Output: & \sent{the man slowly}
\end{tabular}
\end{table}

During training, the word to delete is selected in the same way as selecting the word to replace in a seed sentence,
i.e. nouns or pronouns at the beginning of a sentence.
At test time, the smoother is expected to fill in words to connect the topic word with the seed sentence in a grammatical way,
e.g., \nl{\underline{the negotiator is} just a woman trying to peace her life back together.}
(the part rewritten by the smoother is underlined).

\begin{table*}[t]
    \centering
    \small{
    \begin{tabular}{L{1.5cm}L{7.5cm}rrrr}
        \toprule
        \multirow{2}{*}{Type} & \multirow{2}{*}{Example} & \multicolumn{2}{c}{\ours} & \multicolumn{2}{c}{\kao}    \\
                                                           \cmidrule(lr){3-4}          \cmidrule(lr){5-6}          
                              &                          & Count & Funniness & Count & Funniness \\
        \midrule
        Pun & Yesterday a cow saved my life---it was \underline{bovine} intervention. & 33 & 1.13 & 141 & 1.09 \\
        Swap-pun & Yesterday a cow saved my life---it was \underline{divine} intervention. & 33 & 0.05 & 0 & --- \\
        Non-pun & The workers are all involved in studying the spread of \underline{bovine} TB. & 64 & -0.34 & 257 & -0.53 \\ 
        \bottomrule
    \end{tabular}
    \caption{Dataset statistics and funniness ratings of \ours and \kao.
        Pun or alternative words are underlined in the example sentence.
        Each worker's ratings are standardized to $z$-scores.
        There is clear separation among the three types in terms of funniness,
        where pun $>$ swap-pun $>$ non-pun.
    }
    \label{tab:funniness_data}
    }
\end{table*}

\begin{table*}[t]
    \newcommand{\pv}{$p$-value}
    \centering
    \small{
    \begin{tabular}{lrcrcrcrcrc}
        \toprule
        \multirow{2}{*}{Metric} & \multicolumn{4}{c}{Pun and non-pun} & \multicolumn{2}{c}{Pun and swap-pun} & \multicolumn{4}{c}{Pun} \\
                                  \cmidrule(lr){2-5} \cmidrule(lr){6-7}  \cmidrule(lr){8-11}
                                & \multicolumn{2}{c}{\ours} & \multicolumn{2}{c}{\kao} & \multicolumn{2}{c}{\ours} & \multicolumn{2}{c}{\ours} & \multicolumn{2}{c}{\kao} \\
        \midrule
        Surprisal (\sratio) & \bf 0.46  & $p$=0.00 & \bf 0.58 & $p$=0.00 & \bf 0.48 & $p$=0.00 &     0.26 & $p$=0.15 &     0.08 & $p$=0.37  \\
        Ambiguity       & \bf 0.40  & $p$=0.00 & \bf 0.59 & $p$=0.00 &     0.18 & $p$=0.15 &     0.00 & $p$=0.98 &     0.00 & $p$=0.95  \\
        Distinctiveness &    -0.17  & $p$=0.10 & \bf 0.29 & $p$=0.00 &     0.15 & $p$=0.24 & \bf 0.41 & $p$=0.02 & \bf 0.27 & $p$=0.00  \\
        Unusualness     & \bf 0.37  & $p$=0.00 & \bf 0.36 & $p$=0.00 &     0.19 & $p$=0.12 &     0.20 & $p$=0.27 &     0.11 & $p$=0.18  \\
        \bottomrule
    \end{tabular}
    \caption{Spearman correlation between different metrics and human ratings of funniness.
        Statistically significant correlations with \pv{} $<0.05$ are bolded.
        Our surprisal principle successfully differentiates puns from non-puns and swap-puns.
        Distinctiveness is the only metric that correlates strongly with human ratings within puns.
        However, no single metric works well across different types of sentences.
    }
    \vspace{-1em}
    \label{tab:correlation}
    }
\end{table*}

\section{Experiments}
\label{sec:experiments}
We first evaluate how well our surprisal principle predicts the funniness of sentences perceived by humans (\refsec{principle_analysis}),
and then compare our pun generation system and its variations
with a simple retrieval baseline and a neural generation model~\cite{yu2018neural} (\refsec{results}).
We show that the local-global surprisal scores strongly correlate with human ratings of funniness,
and all of our systems outperform the baselines based on human evaluation.
In particular, \retrieveSwapTopic (henceforth \system) achieves the highest success rate and average funniness score among all systems.

\subsection{Datasets}
\label{sec:data}
We use the pun dataset from
2017 SemEval task7~\cite{semeval2017pun}.
The dataset contains 1099 human-written puns annotated with pun words and alternative words,
from which we take 219 for development.
We use BookCorpus~\cite{zhu2015moviebook} as the generic corpus for retrieval
and training various components of our system.

\subsection{Analysis of the Surprisal Principle}
\label{sec:principle_analysis}
We evaluate the surprisal principle by analyzing how well the local-global surprisal score (Equation \refeqn{ratio}) predicts funniness rated by humans.
We first give a brief overview of previous computational accounts of humor,
and then analyze the correlation between each metric and human ratings.

\paragraph{Prior funniness metrics.}
\citet{kao2015pun} proposed two information-theoretic metrics:
\emph{ambiguity} of meanings and \emph{distinctiveness} of supporting words.
Ambiguity says that the sentence should support both the pun meaning and the alternative meaning.
Distinctiveness further requires that the two meanings be supported by distinct sets of words.

In contrast, our metric based on the surprisal principle imposes additional requirements.
First, surprisal says that while both meanings are acceptable (indicating ambiguity),
the pun meaning is unexpected based on the local context.
Second, the local-global surprisal contrast requires the pun word to be well supported in the global context. 

Given the anomalous nature of puns,
we also consider a metric for \emph{unusualness} based on normalized log-probabilities under a language model~\cite{pauls2012treelets}:
\begin{align}
  \text{Unusualness} \!\eqdef\! - \frac{1}{n}\! \log \!\p{\! p(x_1,\ldots,x_n) /\! \prod_{i=1}^n p(x_i) \!}\!.
\end{align}

\paragraph{Implementation details.}
Both ambiguity and distinctiveness are based on
a generative model of puns.
Each sentence has a latent variable $z \in \{\pw, \aw\}$ corresponding to the pun meaning and the alternative meaning. 
Each word also has a latent meaning assignment variable $f$
controlling whether it is generated from an unconditional unigram language model
or a unigram model conditioned on $z$.
Ambiguity is defined as the entropy of the posterior distribution over $z$ given all the words,
and distinctiveness is defined as the symmetrized KL-divergence between distributions
of the assignment variables given the pun meaning and the alternative meaning respectively.
The generative model relies on $p(x_i \mid z)$,
which \citet{kao2015pun} estimates using human ratings of word relatedness.
We instead use the skip-gram model described in \refsec{pungen} as we are interested in a fully-automated system.

For local-global surprisal and unusualness, we estimate probabilities of text spans using a neural language model trained on WikiText-103~\cite{merity2016pointer}.\footnote{\url{https://dl.fbaipublicfiles.com/fairseq/models/wiki103_fconv_lm.tar.bz2}.}
The local context window size ($d$ in Equation \refeqn{local_global}) is set to 2.

\paragraph{Human ratings of funniness.}
Similar to \citet{kao2015pun},
to test whether a metric can differentiate puns from normal sentences,
we collected ratings for both
puns from the SemEval dataset
and \emph{non-puns} retrieved from the generic corpus containing either \pw or \aw.
To test the importance of surprisal,
we also included \emph{swap-puns} where \pw is replaced by \aw, which results in sentences that are ambiguous but not necessarily surprising.

We collected all of our human ratings on Amazon Mechanical Turk (AMT).
Workers are asked to answer the question \nl{How funny is this sentence?} on a scale from 1 (not at all) to 7 (extremely).
We obtained funniness ratings
on 130 sentences from the development set with 33 puns, 33 swap-puns, and 64 non-puns. 
48 workers each read roughly 10--20 sentences in random order, counterbalanced for sentence types of non-puns, swap-puns, and puns.
Each sentence is rated by 5 workers,
and we removed 10 workers whose maximum Spearman correlation with other people rating the same sentence is lower than 0.2. 
The average Spearman correlation among all the remaining workers (which captures inter-annotator agreement) is 0.3. 
We $z$-scored the ratings of each worker for calibration and took the average $z$-scored ratings of a sentence as its funniness score.

\reftab{funniness_data} shows the statistics of our annotated dataset (\ours) and \citet{kao2015pun}'s dataset (\kao).
Note that the two datasets have different numbers and types of sentences, and the human ratings were collected separately. 
As expected, puns are funnier than both swap-puns and non-puns. Swap-puns are funnier than non-puns, possibly because they have inherit ambiguity brought by the \retrieveSwap operation.

\paragraph{Automatic metrics of funniness.}
We analyze the following metrics:
local-global surprisal (\sratio),
ambiguity, distinctiveness, and unusualness, with respect to their correlation with human ratings of funniness.
For each metric, we standardized the scores and outliers beyond two standard deviations are set to $+2$ or $-2$ accordingly.\footnote{Since both \sratio and distinctiveness are unbounded, bounding the values gives more reliable correlation results.
}
We then compute the metrics' Spearman correlation with human ratings.
On \kao, we directly took the ambiguity scores and distinctiveness scores from the original implementation which requires human-annotated word relatedness.\footnote{\url{https://github.com/amoudgl/pun-model}}
On \ours, we used our reimplemention of \citet{kao2015pun}'s algorithm but with the skip-gram model.

The results are shown in \reftab{correlation}.
For puns and non-puns, all metrics correlate strongly with human scores,
indicating all of them are useful for pun detection.
For puns and swap-puns, only local-global surprisal (\sratio) has strong correlation,
which shows that
surprisal is important for characterizing puns.
Ambiguity and distinctiveness do not differentiate pun word from the alternative word,
and unusualness only considers probability of the sentence with the pun word,
thus they do not correlate as significantly as \sratio.

Within puns, only distinctiveness has significant correlation,
whereas the other metrics are not fine-grained enough to differentiate good puns from mediocre ones.
Overall, no single metric is robust enough to score funniness across all types of sentences,
which makes it hard to generate puns by optimizing automatic metrics of funniness directly.

There is slight inconsistency between results on \ours and \kao. Specifically, for puns and non-puns, the distinctiveness metric shows a significant correlation with human ratings on \kao but not on \ours. 
We hypothesize that it is mainly due to differences in the two corpora and noise from the skip-gram approximation.
For example, our dataset contains longer sentences with an average length of 20 words versus 11 words for \kao.
Further, \citet{kao2015pun} used human annotation of word relatedness while we used the skip-gram model to estimate $p(x_i \mid z)$. 

\begin{table}[t]
    \centering
    \small{
    \begin{tabular}{p{1.6cm}rrr}
         \toprule
         Method & Success  &  Funniness & Grammar \\ \midrule
         NJD & 9.2\% & 1.4 & 2.6 \\
         R & 4.6\% & 1.3 & \bf 3.9 \\
         R+S & 27.0\% & 1.6 & 3.5 \\
         R+S+T+M & 28.8\% & \bf 1.7 & 2.9 \\ 
         \system & \bf 31.4\% & \bf 1.7 & 3.0 \\
         \midrule
         Human & 78.9\% & 3.0 & 3.8 \\ \bottomrule
         
    \end{tabular}
    }
    \caption{Human evaluation results of all systems. We show average scores of funniness and grammaticality on a 1-5 scale and success rate computed from yes/no responses. We compare with two baselines: \pku (NJD) and \retrieve (R). R+S, \system, and R+S+T+M are three variations of our method: \retrieveSwap,  \retrieveSwapTopic, and \retrieveSwapTopicCombiner, respectively. Overall, \system performs the best across the board.
    }
    \label{tab:results_score}
    \vspace{-1em}
\end{table}

\subsection{Pun Generation Results}
\label{sec:results}
\paragraph{Systems.}
We compare with a recent neural pun generator~\citep{yu2018neural}.
They proposed an unsupervised approach based on generic language models to generate homographic puns.\footnote{Sentences where the pun word and alternative word have the same written form (e.g., \emph{bat}) but different senses.}
Their approach takes as input two senses of a target word (e.g., \emph{bat.n01}, \emph{bat.n02} from WordNet synsets),
and decodes from both senses jointly by taking a product of the probabilities conditioned on the two senses respectively (e.g., \emph{bat.n01} and \emph{bat.n02}),
so that both senses are reflected in the output. 
To ensure that the target word appears in the middle of a sentence, they decode backward from the target word towards the beginning and then decode forward to complete the sentence.  
We adapted their method to generate homophonic puns by considering \pw and \aw as two input senses
and decoding from the pun word. We retrained their forward / backward language models on the same BookCorpus used for our system.
For comparison, we chose their best model (\pku), which mainly captures ambiguity in puns.

In addition, we include a retrieval baseline (\retrieve)
which simply retrieves sentences containing the pun word.

For our systems, we include the entire progression of methods described in~\refsec{approach} (\retrieveSwap, \retrieveSwapTopic, and \retrieveSwapTopicCombiner).

\paragraph{Implementation details.}
The key components of our systems include a retriever,
a skip-gram model for topic
word prediction, a type consistency checker,
and a neural smoother.
Given an alternative word, the retriever returned 500 candidates,
among which we took the top 100 as seed sentences (\refsec{pungen} local surprisal).
For topic words,
we took the top 100 words predicted by the skip-gram model
and filtered them to ensure type consistency with the deleted word (\refsec{pungen} global surprisal).
The WordNet path similarity threshold for type consistency was set to 0.3.

The skip-gram model was trained on BookCorpus with $d_1\!\!=\!\!5$ and $d_2\!\!=\!\!10$ in Equation~\refeqn{skipgram}.
We set the word embedding size to 300
and trained for 15 epochs using Adam~\cite{kingma2014adam} with a learning rate of 0.0001. 
For the neural smoother, we trained a single-layer LSTM (512 hidden units) \sts model with attention on BookCorpus.
The model was trained for 50 epochs using AdaGrad~\cite{duchi10adagrad} with a learning rate of 0.01 and a dropout rate of 0.1.

\begin{table}[t]
    \centering
    \small{
    \begin{tabular}{l@{ }|rr|rr}
         \toprule
         \multirow{2}{*}{Aspect} & \multicolumn{2}{c|}{\system v.s. NJD} & \multicolumn{2}{c}{\system v.s. Human} \\ 
         & win \% & lose \% & win \% & lose \% \\ \midrule
         Success & 48.0 & 5.3 & 6.0 & 78.7 \\
         Funniness & 56.7 & 25.3 & 10.7 & 85.3 \\
         Grammar & 60.7 & 30.0 & 8.0 & 82.0 \\ \bottomrule
        
    \end{tabular}
    }
    \caption{Pairwise comparison between \system and \pku (NJD), and between \system and human written puns.
    Win \% (lose \%) is the percentage among the human-rated 150 sentences where \system achieves a higher (lower) average score compared to the other method. The rest are ties.}
    \label{tab:results_paired}
    \vspace{-1em}
\end{table}

\begin{table*}[h!]  
  \centering
  \small{
    \begin{tabular}{l|p{13cm}|l}
    \toprule
    Method & Example & Rating \\ \hline
    \multicolumn{2}{l}{\bf 1. Pun/alternative word pair: butter -- better} & \\ \midrule
    NJD & He is going to come up with the butter a `very good' approach to the world's economic crisis, the world's biggest economic climate.  & 1 \\
    \system & Well, gourmet did it, he thought, it'd butter be right.  & 2 \\
    Human & Why did the dairy churn? The less said, the butter...  & 1.5 \\ \midrule
    \multicolumn{2}{l}{\bf 2. Pun/alternative word pair: peace -- piece}  & \\ \midrule
    NJD & Further, he said, at the end of the peace, it's not clear that it will be a good example. & 1 \\ 
    \system & That's because negotiator got my car back to me in one peace.  & 1.5 \\
    Human & Life is a puzzle; look here for the missing peace.  & 3 \\ \midrule
    \multicolumn{2}{l}{\bf 3. Pun/alternative word pair: flour -- flower}  & \\ \midrule
    NJD & Go, and if you are going on the flour.  & 1 \\
    \system & Butter want to know who these two girls are, the new members of the holy flour.  & 1.5 \\
    Human & Betty crocker was a flour child.  & 4.5 \\ \midrule
    \multicolumn{2}{l}{\bf 4. Pun/alternative word pair: wait -- weight}  & \\ \midrule
    NJD & Gordon Brown, Georgia's prime minister, said he did not have to wait, but he was not sure whether he had been killed.  & 0 \\ 
    \system & Even from the outside, I could tell that he'd already lost some wait.  & 2 \\
    Human & Patience is a virtue heavy in wait.  & 3 \\ \bottomrule
    \end{tabular}
    }
    \caption{Examples of generated puns with average human ratings of funniness (1-5). 0 means that all ratings are N/A (does not make sense).}
    \label{tab:casestudy}
    \vspace{-1em}
\end{table*}

\paragraph{Human evaluation.}
We hired workers on AMT to rate
outputs from all 5 systems together with 
expert-written puns from the SemEval pun dataset. 
Each worker was shown a group of sentences generated by all systems (randomly shuffled) given the same pun word and alternative word pair.
Workers were asked to rate each sentence on three aspects: (1) success (\nl{Is the
sentence a pun?}),\footnote{They were shown the definition from \newcite{miller2017semeval}.}
(2) funniness (\nl{How funny is the sentence?}), and (3) grammaticality
(\nl{How grammatical is the sentence?}).
Success was rated as yes/no, and funniness and grammaticality were rated on a scale from 1 (not at all) to 5 (very).
We also included a N/A choice (does not make sense) for funniness to exclude cases where the sentence are not understandable.
Workers were explicitly instructed to try their best to give different scores for sentences in the same group. 

We evaluated 150 pun/alternative word pairs.  Each generated sentence was rated by 5 workers and their scores were averaged.
N/A ratings were excluded unless all ratings of a sentence were N/A, in which case we set its score to 0.
We attracted 65, 93, 66 workers for the success, funniness, and grammaticality surveys respectively, and removed 3, 4, 4 workers because their maximum Spearman correlation with other workers was lower than 0.2. 
We measure inter-annotator agreement using average Spearman correlation among all workers, and the average inter-annotator Spearman correlation for success, funniness, and grammaticality are 0.57, 0.36, and 0.32, respectively. 

\begin{figure}
    \centering
    \includegraphics[scale=0.32]{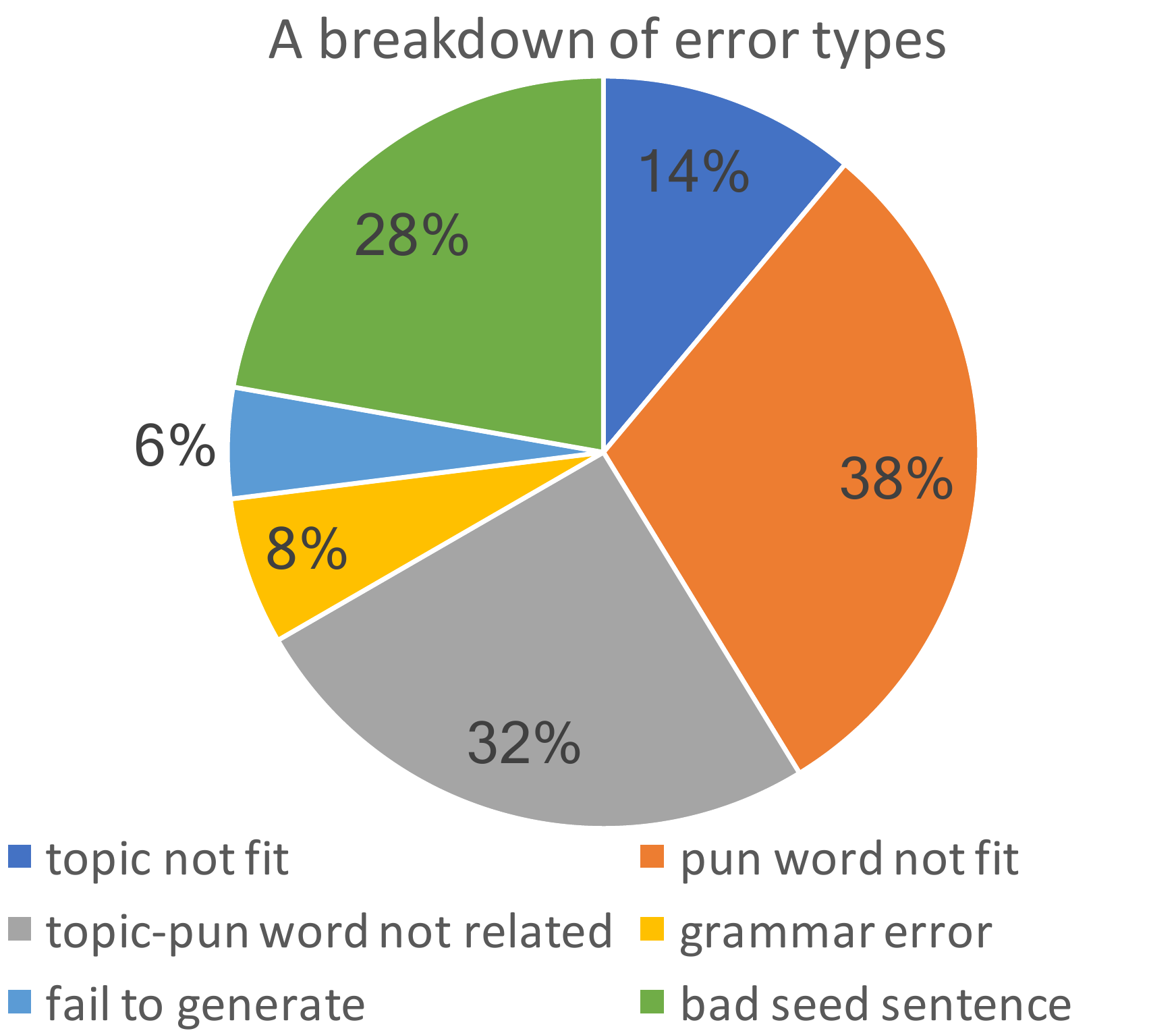}
    \caption{Error case breakdown shows that the main issues lie in finding seed sentences that accommodates both the pun word and the topic word (topic not fit + pun word not fit + bad seed sentence).}
    \label{fig:error}
    \vspace{-1em}
\end{figure}

\reftab{results_score} shows the overall results.
All 3 of our systems outperform the baselines in terms of success rate and funniness.
More edits (i.e. swapping, inserting topic words) made the sentence less grammatical,
but also much more like puns (higher success rate).
Interestingly, introducing the neural smoother did not improve grammaticality and hurt success rate slightly.
Manual inspection shows that ungrammaticality is often caused by improper topic word,
thus fixing its neighboring words does not truly solve the problem.
For example, filling \nl{drum} (related to \nl{lute}) in 
\nl{if that was \blank it was likely that another body would turn up soon, because someone probably wouldn't want to share the lute.}.
In addition, when the neural model is given a rare topic word, it tends to rewrite it to a common phrase instead,
again showing that supervised learning is against the spirit of generating novel content.
For example, inserting \nl{gentlewoman} to \nl{\blank not allow me to ...} produces \nl{these people did not allow me to ...}.
Overall, our \system performs the best and tripled the success rate of \pku with improved funniness and grammaticality scores.
Nevertheless, there is still a significant gap between generated puns and expert-written puns across all aspects, indicating that pun generation remains an open challenge. 

\reftab{results_paired} shows the pairwise comparison results among our best model \system, \pku, and expert-written puns.
Given the outputs of two systems, we decided win/lose/tie by comparing the average scores of both outputs.
We see that \system dominates \pku with $> 50\%$ winning rate on funniness and grammaticality.
On success rate, the two methods have many ties since they both have relatively low success rate.
Our generated puns were rated funnier than expert-written puns around $10\%$ of the time.

\subsection{Error Analysis}
In \reftab{casestudy}, we show example outputs of our \system, the \pku baseline, and expert-written puns.
\system sometimes generates creative puns that are rated even funnier than human-written puns (example 1).
In contrast, \pku at best generates ambiguous sentences (example 2 and 3)
and sometimes the sentences are ungrammatical (example 1) or hard to understand (example 4).
The examples also show the current limitation of \system.
In example 3, it failed to realize that \nl{butter} is not animate thus cannot \nl{want} since our type consistency checker is very simple.

To gain further insights on the limitation of our system, we randomly sampled 50 unsuccessful generations (labeled by workers) to analyze the issues. We characterized the issues into 6 non-exclusive categories:
(1) weak association between the local context and \aw (e.g., \nl{...in the form of a batty (bat)});
(2) \pw does not fit in the local context, often due to different POS tags of \aw and \pw (e.g., \nl{vibrate with a taxed (text)});
(3) the topic word is not related to \pw (e.g., \nl{pagan} vs \nl{fabrication});
(4) the topic word does not fit in its immediate context, often due to inconsistent types (e.g., \nl{slider won't go...}),
(5) grammatical errors;
and (6) fail to obtain seed sentences or topic words.
A breakdown of these errors is shown in \reffig{error}.
The main issues lie in finding seed sentences that accommodate both the pun word and the topic word.
There is also room for improvement in predicting pun-related topic words.

\section{Discussion and Related Work}
\label{sec:disccussion}
\subsection{Humor Theory}
Humor involves complex cognitive activities
and many theories attempt to explain what might be considered humorous.
Among the leading theories, the incongruity theory~\cite{veale2004incongruity} is most related to our surprisal principle.
The incongruity theory posits that humor is perceived at the moment of resolving the incongruity between two concepts,
often involving unexpected shifts in perspectives. 
\newcite{ginzburg2015understanding} applied the incongruity theory to explain laughter in dialogues. 
Prior work~\cite{kao2015pun} on formalizing incongruity theory for puns
focuses on ambiguity between two concepts and the heterogeneity nature of the ambiguity.
Our surprisal principle further formalizes unexpectedness (local surprisal) and incongruity resolution (global association).

The surprisal principle is also related to studies in psycholinguistics on the relation between surprisal and human comprehension~\cite{levy2013memory,levy2013surprisal}. Our study suggests it could be a fruitful direction to formally study the relationship between human perception of surprisal and humor.

\subsection{Humor generation}
Early approaches to joke generation~\cite{binsted1996jape,ritchie2005computational} largely rely on templates for specific types of puns.
For example, JAPE~\cite{binsted1996jape} generates noun phrase puns as question-answer pairs,
e.g., \nl{What do you call a [murderer] with [fiber]? A [cereal] [killer].}
\citet{petrovic2013unsupervised} fill in a joke template based on word similarity and uncommonness.
Similar to our editing approach,
\citet{valitutti2013adult} substitutes a word with a taboo word based on form similarity and local coherence to generate adult jokes.
Recently, \citet{yu2018neural} generates puns from a generic neural language model
by simultaneously conditioning on two meanings.
Most of these approaches leverage some assumptions of joke structures, e.g., incongruity, relations between words, and word types. 
Our approach also relies on specific pun structures; we have proposed and operationalized a local-global surprisal principle for pun generation.

\subsection{Creative text generation}
Our work is also built upon generic text generation techniques,
in particular recent neural generation models.
\newcite{hashimoto2018edit} developed a retrieve-and-edit approach
to improve both grammaticality and diversity of the generated text.
\newcite{shen2017style,fu2018style} explored adversarial training to manipulate the style of a sentence.
Our neural smoother is also closely related to \newcite{li2018style}'s
delete-retrieve-edit approach to text style transfer.

Creative generation is more challenging as it requires both formality (e.g., grammaticality, rhythm, and rhyme) and novelty. Therefore, many works (including us) impose strong constraints on the generative process, such as~\newcite{petrovic2013unsupervised,valitutti2013adult} for joke generation, \newcite{ghazvininejad2016poem} for poetry generation, and \newcite{peng2019plan} for storytelling.

\section{Conclusion}
In this paper, we tackled pun generation by developing and exploring a local-global surprisal principle.
We show that a simple instantiation based on only a language model trained on non-humorous text
is effective at detecting puns (though is not fine-grained enough to detect the degree of funniness within puns). 
To generate puns, we operationalize the surprisal principle with a retrieve-and-edit framework to create contrast in the amount of surprise in local and global contexts.
While we improve beyond current techniques, we are still far from human-generated puns.

While we believe the local-global surprisal principle is a useful conceptual tool,
the principle itself is not quite yet formalized in a robust enough way that can be
be used both as a principle for evaluating sentences and can be directly optimized to generate puns.
A big challenge in humor, and more generally, creative text generation,
is to capture the difference between creativity (novel but well-formed material) and nonsense (ill-formed material).
Language models conflate the two, so developing methods that are nuanced enough to recognize this difference
is key to future progress.

\section*{Acknowledgments}
This work was supported by the DARPA CwC program under ISI prime contract no. W911NF-15-1-0543 and ARO prime contract no. W911NF-15-1-0462.
We thank Abhinav Moudgil and Justine Kao for sharing their data and results. 
We also thank members of the Stanford NLP group and USC Plus Lab for insightful discussions.

\section*{Reproduciblility}
All code, data, and experiments for this paper are available on the CodaLab platform:
\url{https://worksheets.codalab.org/worksheets/0x5a7d0fe35b144ad68998d74891a31ed6
}.


\pl{don't forget to proofread and do a spell check!}

\clearpage
\bibliographystyle{acl_natbib}
\bibliography{refdb/all}


\end{document}